\documentclass[conference]{IEEEtran}
\IEEEoverridecommandlockouts

\usepackage{comment}
\renewcommand{\baselinestretch}{0.92}
\usepackage{changepage}
\usepackage{amsmath}
\usepackage{amssymb}
\usepackage[dvips]{graphicx}
\usepackage{setspace}
\usepackage{epsfig}
\usepackage{color}
\usepackage{morefloats}
\usepackage{enumerate}
\usepackage{bbm}
\usepackage{cases}
\usepackage{enumitem}

\usepackage{graphicx}

\usepackage{csquotes}

\allowdisplaybreaks

\usepackage{caption}
\usepackage{subcaption}
\usepackage{amsmath}
\usepackage{multirow}

\usepackage[letterpaper,
            left=0.64in,
            right=0.64in,
            top=0.7in,
            bottom=1.08in]{geometry}

\usepackage{stfloats}
\usepackage{booktabs}

\begin{document}
\renewcommand{\baselinestretch}{0.85}

\title{PRB-RUPFormer: A Recursive Unified Probabilistic Transformer for Residual PRB Forecasting}
\author{Saad Masrur\vspace{-0.5cm}}
\author{Saad Masrur$^{\star,\dag}$, 
Yuxuan Jiang$^{\star}$, 
Matti Hiltunen$^{\star}$, 
Ajay Rajkumar$^{\star}$, and 
\.{I}smail G\"{u}ven\c{c}$^{\dag}$\\
$^{\star}$AT\&T RAN Technology, Bedminster, NJ\\
$^{\dag}$Department of Electrical and Computer Engineering, North Carolina State University, Raleigh, NC\\
 {\tt \{smasrur,iguvenc\}@ncsu.edu}
}


\vspace{-0.7cm}
\maketitle
\thispagestyle{empty}

\begin{abstract}
\pagestyle{empty}
Accurate forecasting of residual Physical Resource Blocks (PRBs) is critical for proactive network slice provisioning, energy-efficient operation, and spectrum-aware decision making in cellular systems, where residual PRBs serve as a practical proxy for short- and medium-term spectrum availability. Existing PRB prediction methods typically rely only on historical PRB values and are trained independently per carrier or sector, limiting their ability to capture cross-carrier dependencies and providing no measure of forecast uncertainty. Moreover, point forecasts alone are insufficient for robust spectrum-aware control under highly variable traffic conditions. This paper proposes \texttt{PRB-RUPFormer}, a recursive unified probabilistic Transformer for residual PRB forecasting. The proposed model jointly processes multivariate KPI time series using temporal, seasonal, and carrier-aware embeddings, preserving inter-metric temporal coupling during recursive rollout and stabilizing long-horizon forecasting. A single shared model is trained across all carriers and sectors of an eNB, enabling efficient learning of joint traffic dynamics with low computational overhead. Forecast uncertainty is captured through quantile-based prediction intervals, providing confidence-aware estimates of future PRB availability. Evaluations on six months of commercial LTE network data from multiple U.S. locations demonstrate median MAE below $0.05$ and hit probabilities above $0.80$ for both one-day and seven-day recursive forecasts. These probabilistic predictions directly support spectrum-aware RAN functions such as dynamic carrier activation, congestion avoidance, and proactive spectrum sharing, making the proposed framework well-suited for dynamic spectrum access scenarios.

\pagestyle{empty}

\textit{Index~Terms}--- O-RAN, PRB forecasting, Dynamic spectrum access, Transformer models, Probabilistic prediction, RAN intelligence.
\end{abstract}




\vspace{-0.2cm}
\section{Introduction}\label{sec:intro}
\pagestyle{empty}

Accurate forecasting of residual physical resource blocks (PRBs) is critical for a wide range of radio access network (RAN) functions, including network slice provisioning, spectrum management, quality-of-service (QoS) assurance, and energy-efficient operation. Traffic patterns in operational cellular networks exhibit strong temporal variability driven by diurnal usage cycles, user mobility, and event-induced surges. These dynamics directly affect PRB availability across carriers and sectors, and anticipating future PRB utilization enables proactive control actions that mitigate congestion and improve resource utilization.

\begin{figure}
	\includegraphics[width=0.95\linewidth]{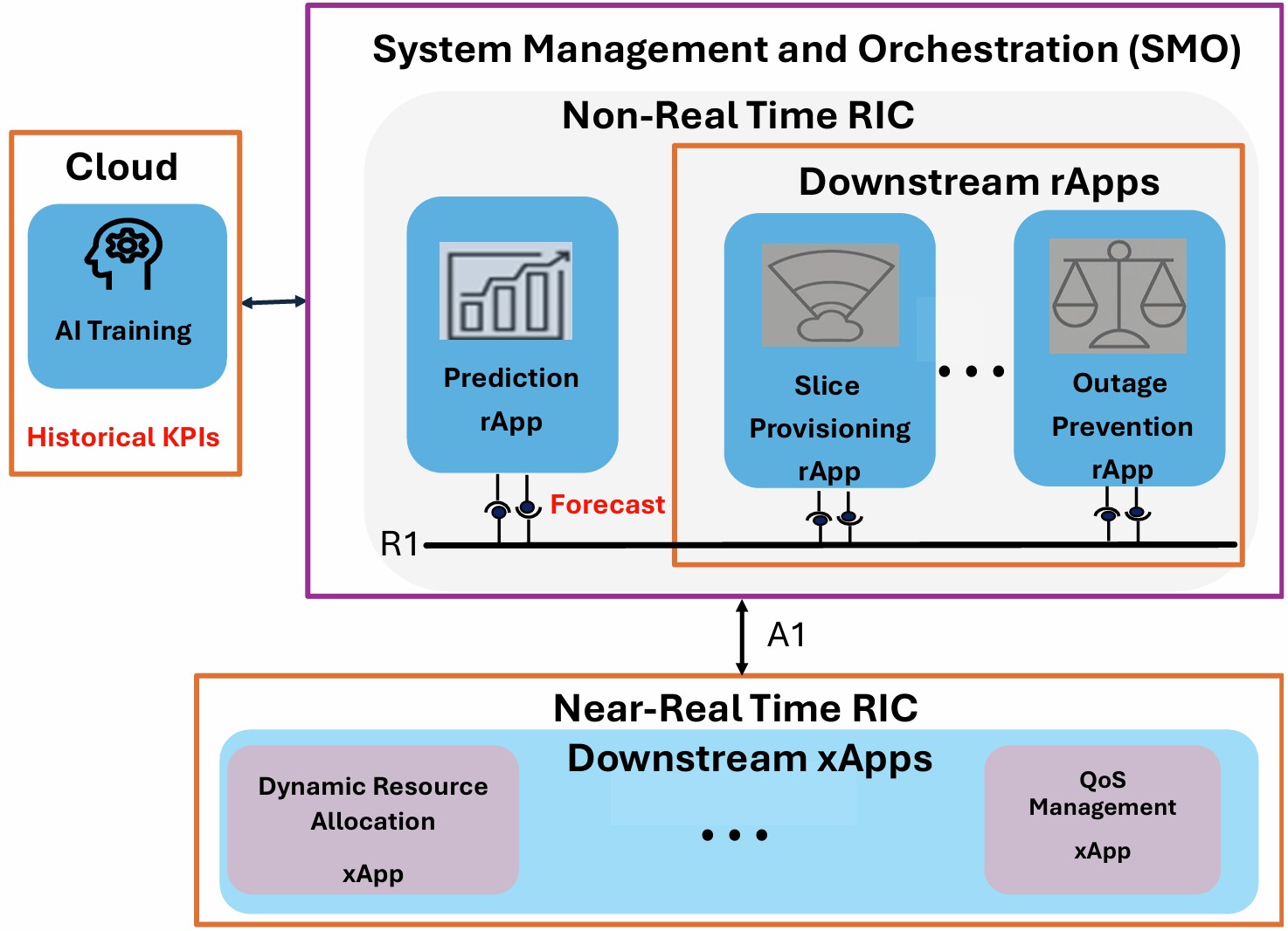}
	\centering
	\caption{Integration of the proposed forecasting model into the O-RAN architecture, showing how probabilistic PRB predictions support rApp and xApp control functions. }
	\label{ORAMimple}
\vspace{-0.5cm}
\end{figure} 


From a spectrum management perspective, residual PRBs provide a fine-grained representation of short- and near-term spectrum availability at the carrier level. Since PRBs correspond to time-frequency resources scheduled at the physical layer, their residual count reflects the spectrum that may be dynamically reallocated, shared, or opportunistically reused. Forecasting residual PRBs, therefore, enables spectrum-aware decisions such as dynamic carrier activation, inter-carrier load balancing, and adaptive spectrum allocation across sectors or network slices. In emerging dynamic spectrum access (DSA) scenarios \cite{7516641}, where spectrum resources are increasingly shared across services and operators, anticipating future PRB availability is essential to reduce interference, prevent congestion, and improve spectral efficiency. By forecasting residual PRBs as a proxy for short- and medium-term spectrum availability, the proposed framework directly supports dynamic spectrum access and spectrum-sharing use cases envisioned in next-generation cellular systems.


These spectrum-aware forecasting capabilities naturally align with the Open Radio Access Network (O-RAN) architecture, which introduces an open, programmable, and disaggregated framework for data-driven RAN control and automation \cite{10024837}. O-RAN defines two types of RAN automation applications distinguished by the order of magnitude of their control-loop latency. rApps operate at the second level and are deployed in the non-real-time RAN Intelligent Controller (RIC), whereas xApps operate at the millisecond level and are deployed in the near-real-time RIC. Both rApps and xApps require timely awareness of future network and spectrum conditions to enable predictive, closed-loop control. As shown in Fig.~\ref{ORAMimple}, the proposed probabilistic PRB forecasting module can be deployed as an O-RAN rApp to provide future network condition awareness for advanced network automation, including network slicing. Specifically, historical KPIs are processed in the cloud to train the forecasting model, which is deployed as a prediction rApp in the non-real-time RIC. The resulting forecasts can be shared with other rApps via the R1 interface and communicated to near-real-time xApps through the A1 interface, enabling spectrum-aware control actions such as dynamic resource allocation, congestion avoidance, and proactive spectrum sharing. While this work focuses on forecasting methodology rather than O-RAN implementation details, the figure highlights the practical relevance of probabilistic PRB prediction for RIC-driven spectrum and resource management.


To address the limitations of existing techniques, this paper introduces the recursive unified probabilistic Transformer (\texttt{PRB-RUPFormer}), a data-driven and computationally efficient Transformer-based architecture for residual PRB forecasting. Unlike conventional per-carrier or per-cell predictors, \texttt{PRB-RUPFormer} employs a single unified model that jointly forecasts residual PRBs across all carriers and sectors of an eNB, enabling it to capture inter-carrier and inter-sector dependencies that naturally arise from shared spectrum usage, carrier aggregation, and coordinated scheduling. This unified design significantly reduces model maintenance and training overhead in large-scale deployments, while improving generalization under data sparsity. The model integrates lightweight attention blocks with temporal and seasonal positional embeddings to efficiently learn long-range traffic dynamics, and performs recursive multi-step forecasting over horizons ranging from minutes to several days. By producing probabilistic forecasts with calibrated confidence intervals, \texttt{PRB-RUPFormer} provides reliable, uncertainty-aware predictions that are well-suited for spectrum-aware and closed-loop RAN control. The main contributions of this paper are summarized as follows: 


\begin{itemize}
    \item A unified, data-driven Transformer architecture that forecasts residual PRBs for all carriers and sectors of an eNB using a single model, capturing inter-carrier and inter-sector dependencies that are missed by per-carrier predictors.
    \item An efficient recursive forecasting strategy that extends short-horizon predictions to multi-day horizons without retraining separate models.
    \item Probabilistic residual PRB forecasts with calibrated confidence intervals, enabling uncertainty-aware spectrum and RAN control decisions.
    \item An extensive evaluation using commercial LTE network data across multiple locations, traffic conditions, and forecasting horizons.
\end{itemize}

The remainder of this paper is organized as follows. Section II reviews related work on PRB and traffic forecasting. Section III presents the system model and problem formulation. Section IV describes the proposed \texttt{PRB-RUPFormer} architecture and training methodology. Section V outlines the experimental setup and evaluation metrics. Numerical results are discussed in Section VI, followed by conclusions in Section VII.

\section{Related Work}

Early work on PRB forecasting primarily relied on statistical and autoregressive models that capture temporal correlations in PRB usage using historical observations alone \cite{premsankar2021data}. While effective for modeling short-term trends, these approaches assume that future PRB utilization depends solely on past PRB values and do not incorporate the influence of traffic load, scheduling dynamics, user mobility, or control-plane activity. As a result, their ability to generalize under non-stationary network conditions is limited.

To capture nonlinear dependencies in cellular networks, machine learning techniques have been applied to traffic and resource forecasting tasks \cite{jiang2023network, mostafa2022downlink, minovski2021throughput}. For example, hybrid models combining KPI selection, neural regression, and seasonal ARIMA have been proposed for downlink throughput forecasting using per-cell historical data \cite{mostafa2022downlink}. Other studies evaluate ensemble learning and neural models to identify KPIs that influence short-term throughput prediction accuracy \cite{minovski2021throughput}. However, these approaches focus on throughput rather than PRB availability, operate independently per cell, and produce point estimates that do not quantify prediction uncertainty, limiting their applicability to spectrum-aware control and admission decisions.

Recurrent neural networks, including LSTM and GRU architectures, have been widely used for cellular traffic forecasting. LSTM-based models have been applied to short-horizon bearer-level throughput and PRB demand prediction using scheduler-level information \cite{gutterman2019ran}, as well as to millisecond-scale PRB utilization forecasting using large LTE datasets \cite{nagib2021deep}. While these methods capture temporal correlations, recurrent structures struggle with long-range dependencies and typically operate on a per-cell or per-carrier basis. Moreover, they generate deterministic point predictions and do not model inter-carrier or inter-sector coupling, which is critical for coordinated spectrum management.

More recently, transformer-based architectures have gained attention for network traffic forecasting due to their ability to model long-range temporal dependencies via self-attention mechanisms \cite{kasuluru2023use}. Existing transformer-based studies in cellular networks, however, primarily target single-step or short-horizon forecasting and are often designed for individual cells or carriers. In addition, most of these approaches provide point estimates only, without uncertainty quantification, which restricts their use in spectrum sharing, slice admission control, and energy-aware RAN optimization where robustness is essential.

From a spectrum management perspective, prior work has highlighted the importance of forecasting resource demand to support proactive allocation and avoid over- or under-provisioning, particularly in the context of RAN slicing and shared spectrum environments \cite{gutterman2019ran, farajzadeh2025data}. However, existing solutions typically treat PRB demand prediction as a supporting component rather than a unified forecasting problem across carriers and sectors, and they do not explicitly address probabilistic forecasting of residual PRBs as a proxy for future spectrum availability.

Within the O-RAN ecosystem, forecasting is widely recognized as a key enabler for intelligent rApps and xApps supporting functions such as PRB utilization prediction, interference mitigation, and proactive mobility management. Despite this recognition, the integration of probabilistic, multi-carrier PRB forecasting into O-RAN control workflows remains largely unexplored. This work addresses this gap by proposing a unified probabilistic Transformer-based architecture for residual PRB forecasting and outlining its deployment within O-RAN-compliant control loops, enabling spectrum-aware, predictive RAN management.

\section{System Model}\label{sec:system}

\begin{figure}
	\includegraphics[width=0.8\linewidth]{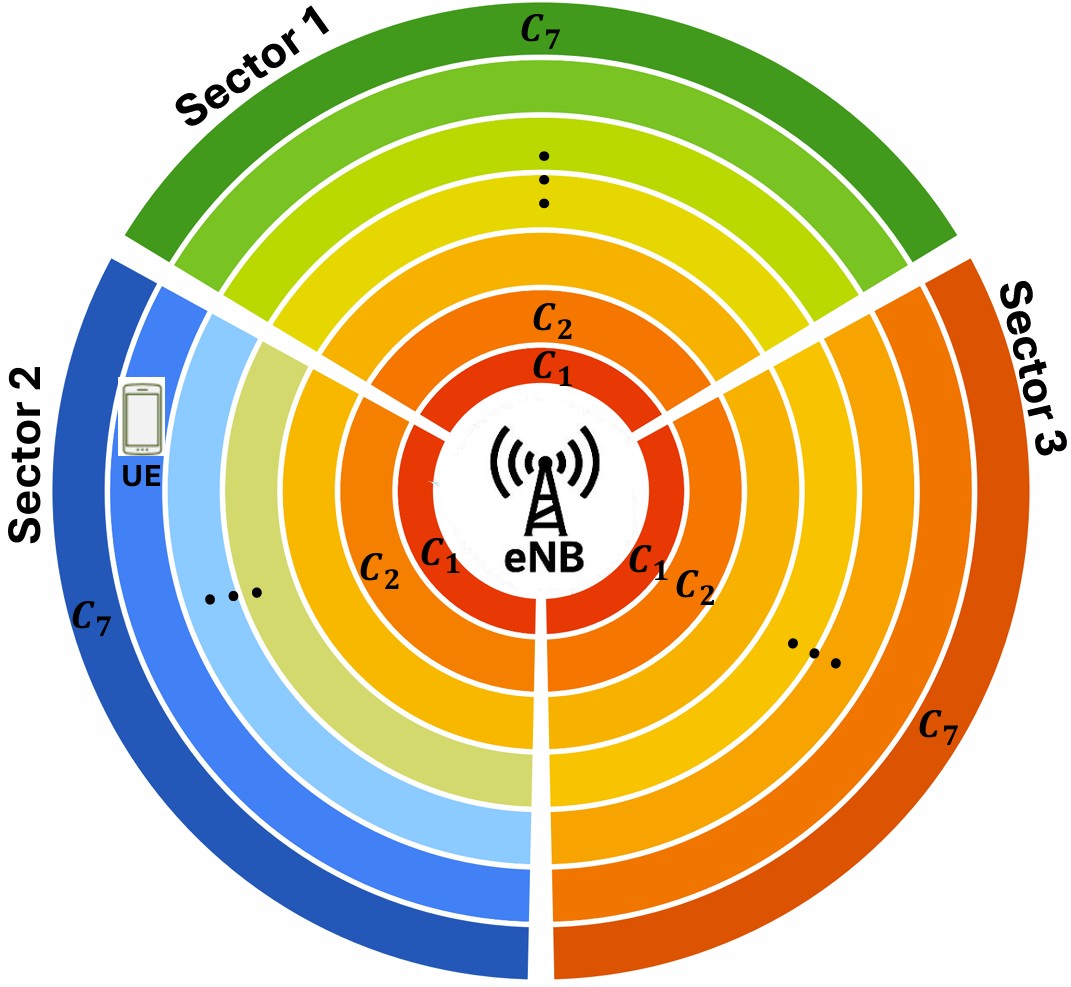}
	\centering
	\caption{Illustration of the LTE eNB configuration used in this study, consisting of three sectors, each operating seven carriers.}
	\label{Sectors}
\vspace{-0.5cm}
\end{figure}

We consider a commercial LTE eNB deployment composed of multiple sectors, typically three sectors covering approximately 120 degrees each. Each sector operates several component carriers with different center frequencies and bandwidths, forming a multi-layer structure. As illustrated in Fig.~\ref{Sectors}, these coverage layers overlap in practice, and their boundaries are not strict; the effective footprint of each carrier varies with path loss, shadowing, antenna configuration, and traffic load. Training separate forecasting models for each carrier, as commonly done in prior studies, does not leverage the inherent coupling across carriers and sectors and leads to redundant model instances. This motivates the use of a unified forecasting framework capable of capturing cross-carrier and cross-sector dependencies, which is essential in dynamic spectrum access scenarios where spectrum resources may be flexibly reassigned, shared, or opportunistically reused across carriers and sectors.

At the eNB, a large number of key performance indicators (KPIs) are measured at the transmission time interval (TTI) level and are commonly aggregated to coarser resolutions for monitoring and analytics in operational cellular networks. In this work, KPIs are aggregated over fixed 15-minute intervals, which is an industry-standard granularity for cell- and carrier-level performance reporting. At each discrete time index \(t\), the network reports multiple KPIs, from which a subset is selected based on Pearson correlation with residual PRB utilization to capture the dominant drivers of PRB consumption and short-term spectrum availability. The forecasting target is the residual PRB ratio, defined for each carrier as:
\begin{equation}
r_t = \frac{N_{\text{PRB}}^{\text{tot}} - N_{\text{PRB}}^{\text{used}}(t)}{N_{\text{PRB}}^{\text{tot}}},
\label{eq:residual_prb}
\end{equation}
where \(N_{\text{PRB}}^{\text{tot}}\) denotes the total number of PRBs available to the carrier and \(N_{\text{PRB}}^{\text{used}}(t)\) is the number of PRBs utilized during the 15-minute interval at time (t). This normalized formulation bounds the residual PRB ratio to the interval ([0,1]) and serves as a practical proxy for short-term spectrum availability, enabling meaningful comparison across carriers and supporting spectrum-aware decision making.

The selected KPI vector at time (t) is defined as:
\begin{align} \small X_t = \big[ &\text{PRB\_MEAN}_t,\; \text{PRB\_TOTAL}_t,\; \text{ACTIVE\_TTI}_t,\; \nonumber \\ &\text{PRB\_PDSCH}_t,\; \text{PRB\_PUCCH}_t,\; \text{UE\_MAX}_t,\; \text{UE\_AVG}_t,\;\nonumber\\ & \text{DL\_TPUT}_t,\; r_t \big].\label{eq:Xt} \end{align}

In words, the vector \(\mathbf{X}_t\) includes KPIs that jointly capture traffic demand, scheduling behavior, and user activity, including mean PRB usage per TTI, total PRBs, active TTIs, PRBs allocated to shared and control channels, the maximum and average number of connected UEs, downlink throughput, and the residual PRB ratio. Together, these features provide the contextual information required for accurate and stable residual PRB forecasting.

Prior work on residual PRB forecasting often relies solely on historical PRB values, implicitly assuming that future PRB availability depends only on its recent history. While this simplifies recursive prediction, it ignores the influence of traffic load, scheduling activity, control signaling, and user dynamics, leading to poor generalization under non-stationary conditions, particularly in DSA scenarios. To address this limitation, we adopt a multivariate formulation in which the model observes the past \(N\) KPI vectors:
\begin{equation}
\mathbf{X}^{\text{inp}}_{t} = \{ X_{t-N+1}, \dots, X_{t} \}~,
\label{eq:input_window}
\end{equation}
and predicts the next \(M\) full KPI vectors, not just the residual PRB. Let
\begin{equation}
\mathbf{Y}^{\text{out}}_{t+1:t+M} = \{ \hat{X}_{t+1}, \dots, \hat{X}_{t+M} \}
\label{eq:future_targets}
\end{equation}
denote the sequence of future KPI vectors. The forecasting problem is defined as learning a mapping
\begin{equation}
\mathbf{f_{\theta}} : \mathbf{X}_{t-N+1:t} \rightarrow \widehat{\mathbf{X}}_{t+1:t+M}~,
\label{eq:forecast_mapping}
\end{equation}
where \( \mathbf{f_{\theta}} \) is the model parameterized by \( \theta \).

\begin{figure*}[t]
	\includegraphics[width=0.75\linewidth]{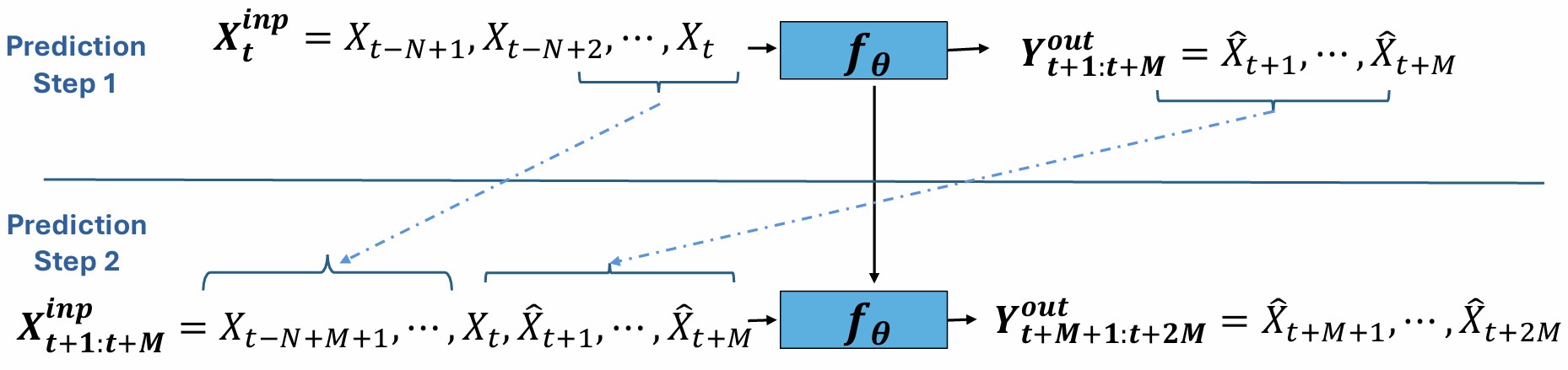}
	\centering
	\caption{Recursive multi-step forecasting where each block of \(M\) predicted KPI vectors is appended to the recent history, and the model \(f_{\theta}\) is reapplied using only the latest \(N\) vectors to maintain a fixed input window.}
	\label{recursion}
\vspace{-0.5cm}
\end{figure*}

This formulation naturally enables recursive multi-step prediction. During inference, the model extends forecasting beyond a single horizon of length \(M\) by feeding predicted KPI vectors back into the input window. After generating the first block of forecasts \(\widehat{\mathbf{X}}_{t+1:t+M}\), the predicted vectors replace the oldest observations while the window length remains fixed at \(N\). The updated window
\begin{equation}
\{ X_{t-N+M+1}, \dots, X_{t},\ \widehat{X}_{t+1}, \dots, \widehat{X}_{t+M} \} ~,
\label{eq:combined_sequence}
\end{equation}
is then fed back into the model to produce the next block of forecasts \(\widehat{\mathbf{X}}_{t+M+1:t+2M}\). This process repeats, with each iteration using the most recent mix of observed and previously predicted KPIs, allowing the prediction horizon to grow without increasing the model input size. As illustrated in fig.~\ref{recursion}, this sliding-window recursion supports long-horizon forecasting by propagating temporal dependencies across both observed and predicted KPI sequences. Joint KPI prediction further preserves inter-metric relationships critical for stable and accurate long-horizon forecasts.

Since future network behavior is uncertain, the model generates probabilistic forecasts
\begin{equation}
p_{\theta}\big( \mathbf{X}_{t+1:t+M} \mid \mathbf{X}_{t-N+1:t} \big)~,
\label{eq:prob_model}
\end{equation}
for just the residual PRB component using quantile-based prediction (see Section V), while the remaining KPIs are predicted deterministically. Although probabilistic outputs are applied only to residual PRBs, the model jointly predicts all KPIs to preserve inter-KPI temporal coupling during recursive rollout. Predicting auxiliary KPIs prevents the accumulation of physically inconsistent states that can arise when residual PRBs are forecast in isolation, thereby stabilizing long-horizon recursion and ensuring that future PRB trajectories remain consistent with underlying traffic load, scheduling activity, and user dynamics. 




\section{Proposed Model: PRB-RUPFormer}
The proposed recursive unified probabilistic Transformer (\texttt{PRB-RUPFormer}) is a unified Transformer-based forecasting architecture designed to predict future KPI vectors and generate probabilistic forecasts for residual PRBs across all carriers of an eNB. The model operates on multivariate KPI sequences, incorporates temporal and categorical embeddings, and supports recursive multi-step prediction using a fixed-length input window. In this section, we elaborate the detailed architecture of the model.

\subsection{Input Embedding Layer}

In this layer, each KPI vector \(X_t \in \mathbb{R}^d\) is first projected into an embedding space through a learnable linear transformation \(W_{\text{proj}} \in \mathbb{R}^{d_\text{emb}}\):

\begin{equation}
    E_t = W_{\text{proj}} \times X_t.
\end{equation}
To provide contextual and temporal structure, the projected vector \(E_t\) is augmented with auxiliary embeddings that encode temporal and categorical information, as described below.

\subsubsection{Positional Embedding \(P_t^{E}\)}
A learned positional embedding that identifies the relative position within the \(N\)-length input window, and helps the Transformer distinguish between early and recent time steps.

\subsubsection{Calendar-Based Temporal Embeddings}

To capture the periodic structure inherent in cellular traffic, the model augments each KPI vector with a set of calendar-based temporal embeddings. These embeddings encode the month, weekday, hour, and minute corresponding to each measurement time, allowing the Transformer to learn seasonal, weekly, and daily load patterns. Formally, each temporal component is implemented as a learned dictionary. The month embedding $(P_t^{M})$ is represented by a $(12 \times d_{\text{emb}})$ lookup table, where each month maps to a learnable vector. The weekday embedding $(P_t^{W})$ uses a $(7 \times d_{\text{emb}})$ table, enabling the model to internalize weekly usage cycles. Hour-of-day information is captured through a $(24 \times d_{\text{emb}})$ embedding matrix $(P_t^{H})$, while minute-level variation (with 15-minute aggregation) is encoded via a $(4 \times d_{\text{emb}})$ table $(P_t^{S})$ corresponding to minute indices $({0,15,30,45})$.

Together, these learned embeddings provide the model with rich temporal context across multiple scales, enabling it to recognize long-term seasonality (e.g., monthly trends), medium-range patterns (e.g., weekday effects), and short-term periodicity (e.g., hourly traffic peaks). This structure enhances the model’s ability to forecast future KPI behavior under varying temporal conditions.

\subsubsection{Carrier Embedding: \(P_t^{C}\)}
Because a single unified model is trained across all carriers of all sectors, we introduce a carrier-identity embedding \(P_t^{C}\) that uses a $(21 \times d_{\text{emb}})$ table that encodes the carrier index associated with each KPI vector. This embedding allows the model to internalize carrier-specific characteristics, such as frequency-dependent coverage ranges, scheduling behavior, and load profiles, while still leveraging shared temporal dynamics learned across carriers.

\subsubsection{Combined Input Embedding}

\begin{figure*}[ht!]
	\includegraphics[width=0.90\linewidth]{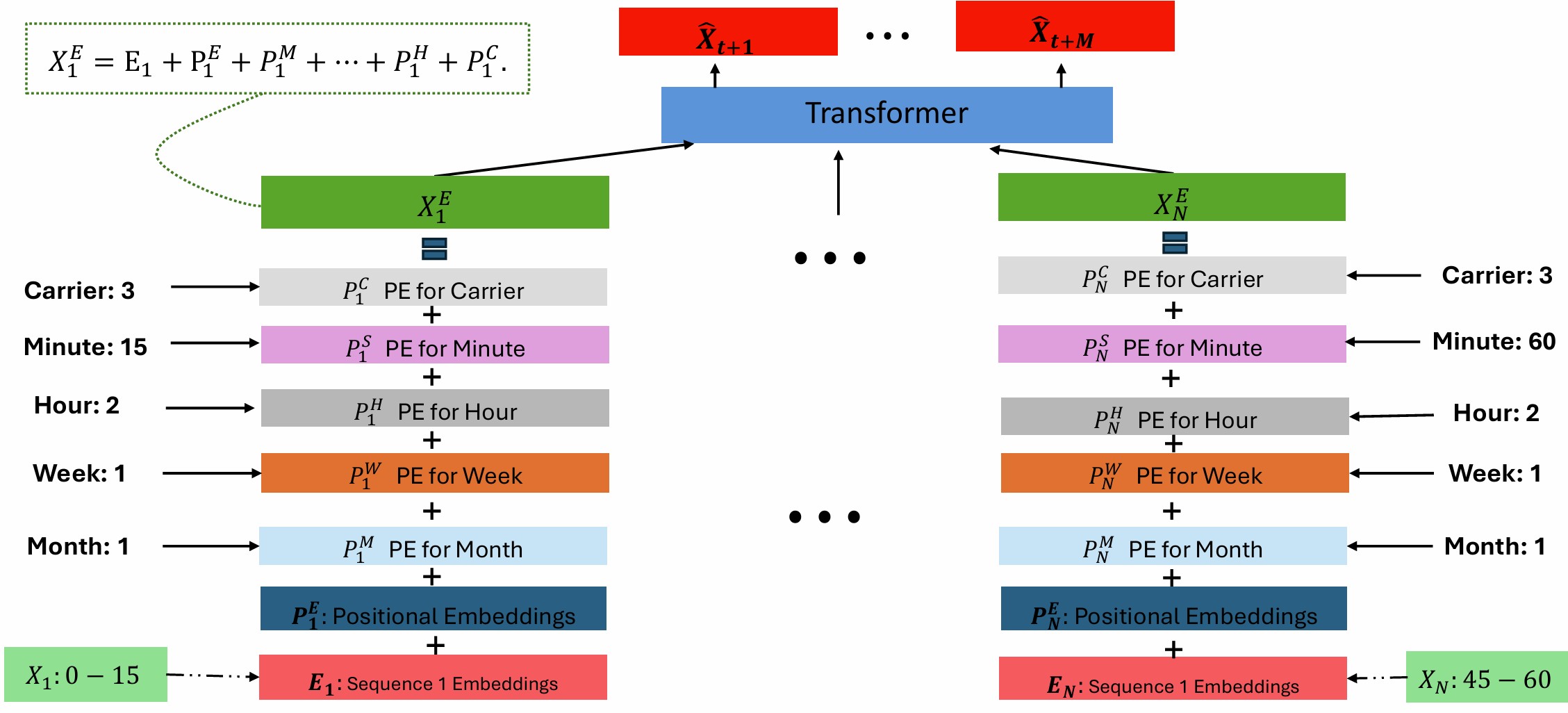}
	\centering
	\caption{The proposed embedding framework for \texttt{PRB-RUPFormer}. }
	\label{embedding_pipeline}
\vspace{-0.35cm}
\end{figure*}

All embeddings are summed to produce the token representation for the Transformer encoder.  The final input representation for time step \(t\) is constructed as:
\begin{equation}
    X_t^{E} = E_t
+P_t^{E}
+ P_t^{M}
+ P_t^{W}
+ P_t^{H}
+ P_t^{S}
+ P_t^{C}~,
\label{embedded}
\end{equation}
where each term corresponds to a different temporal or categorical encoding. The overall embedding pipeline, including positional, temporal, and carrier-specific components, is illustrated in Fig. \ref{embedding_pipeline}.

\subsection{Working Mechanism of the Proposed \texttt{PRB-RUPFormer}}

The proposed \texttt{PRB-RUPFormer} adopts a sequence-to-sequence Transformer architecture designed to model multivariate KPI dynamics and residual PRB uncertainty jointly. The model consists of a Transformer encoder that processes \(N\) historical KPI sequences and a Transformer decoder that autoregressively generates forecasts for the next \(M\) intervals. Although the architecture remains identical during training and inference, the decoder operates under different input conditions, resulting in distinct behaviors.

\subsubsection{Encoder Architecture}
The encoder consists of \(\ell_e\) stacked layers, each containing multi-head self-attention (MHSA) followed by a position-wise feed-forward network, with residual connections and normalization at every sublayer \cite{vaswani2017attention}.
The encoder receives a fixed-length window of \(N\) past embedded input sequnce \({X_{t-N+1}^{E}, \dots, X_t^{E}}\) \eqref{embedded}, and outputs a latent representation sequence  \({Z_{t-N+1}, \dots, Z_t}\) where each vector captures long-range temporal dependencies and cross-KPI interactions. The resulting encoder representations summarize recent network dynamics and are used by the decoder to condition future forecasts.

\subsubsection{Decoder Operation During Training}
 During training, the decoder uses a teacher forcing scheme \cite{sutskever2014sequence}, which stabilizes optimization and enables the model to learn accurate temporal dependencies. At each decoding step, the model is supplied with two types of inputs: (i) future categorical metadata, including the month, weekday, hour, minute, and carrier identity, all of which are known deterministically for future intervals; and (ii) continuous KPI values shifted by one time step, such that the decoder receives the ground truth continuous features from step (k-1) when predicting step (k). For the first prediction step, the continuous inputs are initialized to zero since no earlier ground truth values exist. A causal attention mask ensures that each prediction step only attends to earlier steps within the output horizon. Through cross-attention, the decoder aligns future predictions with the encoder representation \(Z_{t-N+1:t}\), enabling it to integrate both historical context and future time metadata. The decoder prediction head generates
\begin{equation}
\widehat{X}_{t+1:t+M} = f_{\theta}(Z_{t-N+1:t})~,
\end{equation}
where each vector \(\widehat{X}_{t+1:t+M}\) contains: (i) deterministic predictions for all non-PRB KPIs; and (ii) quantile predictions \(\widehat{R}_{t+k,q}\) for residual PRBs with \((q \in {0.1, 0.5, 0.9})\). This hybrid formulation enables probabilistic modeling of PRB availability while keeping the output dimension compact.

\subsubsection{Decoder Operation During Inference}
During inference, ground truth future continuous KPI values are unavailable, so the decoder operates in a fully autoregressive mode. A future metadata sequence containing categorical attributes such as month, weekday, hour, minute, and carrier ID is generated in advance since these values are known deterministically. The corresponding continuous inputs for these future steps are initialized to zero, forming the initial decoder input. The decoder then produces a full block of predictions for the next \(M\) time steps. These predicted continuous values are fed back into the model: the newest \(M\) predictions replace the oldest entries in the encoder input window, while the window length remains fixed at \(N\). The updated window is then re-encoded and passed to the decoder to generate the next block of forecasts \eqref{eq:combined_sequence}.

The encoder-decoder pipeline enables the model to learn both long-term temporal structure from historical windows and short-term fluctuations from localized KPI trends. Deterministic outputs stabilize auxiliary KPI prediction, while quantile-based residual PRB forecasting provides uncertainty estimates essential for slice admission, congestion avoidance, and RAN resource optimization.

\subsubsection{Loss Function and Probabilistic Training Objective}

Future PRB availability is uncertain due to variations in user mobility, traffic bursts, and scheduling decisions. To model this uncertainty, \texttt{PRB-RUPFormer} predicts three quantiles of the residual PRB distribution: \(q = 0.1\), \(0.5\), and \(0.9\). These correspond to pessimistic, median, and optimistic estimates of future residual PRBs. Formally, the model outputs \(\widehat{r}_{t+k,q}\)  which denotes the predicted \(q\)-quantile of the residual PRB at future step \(t+k\). It is defined such that
\begin{equation}
\Pr\big(r_{t+k} \le \widehat{r}_{t+k,q}\big) = q,
\end{equation}
meaning that a fraction \(q\) of the possible future outcomes for the residual PRB is expected to fall below the predicted value.

Each quantile is trained using the pinball (quantile) loss \cite{lim2021temporal}:
\begin{equation}
L_{q}(y, \hat{y}) =
\begin{cases}
q (y - \hat{y}),          & \text{if } y > \hat{y},\\[4pt]
(1 - q) (\hat{y} - y),    & \text{if } y \le \hat{y},
\end{cases}
\label{eq:quantile_loss}
\end{equation}
which penalizes under- and over-prediction asymmetrically to ensure correct quantile behavior. The combined set of quantiles produces calibrated confidence intervals, such as the interval between the tenth and ninetieth percentiles. These intervals quantify uncertainty in future PRB availability and support more reliable slice admission, congestion avoidance, and energy-efficient resource allocation.

For all other KPIs, the model generates deterministic predictions trained using standard L2 loss. Since the paper focuses primarily on accurate residual PRB forecasting, the loss terms are weighted to emphasize quantile prediction. Let \(L_{\text{L2}}\) denote the mean squared error across non-PRB KPIs and \(L_{\text{Q}}\) the sum of quantile losses \eqref{eq:quantile_loss} across ({0.1, 0.5, 0.9}). The total training loss is
\begin{equation}
L_{\text{total}} = \alpha \times L_{\text{L2}} + \beta \times L_{\text{Q}}~,
\label{eq:total_loss}
\end{equation}
where \(\alpha\) and \(\beta\) control the relative emphasis on deterministic KPI prediction and probabilistic residual PRB forecasting. In practice, a larger \(\beta\) is selected to prioritize accurate modeling of residual PRB uncertainty.

For all non-PRB KPIs, \texttt{PRB-RUPFormer} outputs single deterministic values, whereas residual PRBs are predicted through multiple quantiles, a representative value is required for constructing the next input window during recursive inference \eqref{eq:combined_sequence}. For this purpose, the median estimate (the \(q = 0.5\) quantile) is used as the residual PRB value for the subsequent prediction step. The median provides a stable central estimate that avoids optimistic or pessimistic bias.



\section{Numerical Results and Analysis}\label{sec:Nume}

\subsection{Dataset Description and Training Setup}


The evaluation is performed using operational LTE data collected from multiple commercial eNBs deployed across different locations in the United States. The dataset spans approximately six months of measurements and contains KPI records for eNBs configured with three sectors and seven carriers per sector. The first five months of data are used for training the proposed \texttt{PRB-RUPFormer} model, the subsequent 15 days serve as a validation set for hyperparameter tuning, and the final 15 days are held out for testing. This chronological partition ensures that all evaluations reflect true future prediction behavior under realistic network operating conditions.


The proposed \texttt{PRB-RUPFormer} model is implemented in PyTorch and executed on both a GPU platform (NVIDIA Tesla V100-PCIE-32GB) and a high-performance CPU server equipped with a $44$-core, $88$-thread \(x86\_64\) processor operating at up to $3.7$ GHz. This setup enables evaluation of computational efficiency under heterogeneous deployment conditions. \texttt{PRB-RUPFormer} adopts a Transformer encoder–decoder architecture with a model dimensionality of \(d_{\text{emb}} = 64\). The encoder consists of \(\ell_e = 2\) layers, while the decoder contains \(\ell_d = 3\) layers, each employing multi-head attention with eight heads and a feed-forward dimensionality of 256. A dropout rate of 0.1 is applied across all layers. All categorical and temporal embeddings, including positional and calendar-based components, are projected to the same dimensionality and learned jointly with the model parameters.

For all experiments, the input sequence length is fixed to \(N = 4\) historical time steps, corresponding to one hour of aggregated KPI information. The model predicts \(M = 2\) future steps in each decoding cycle. During inference, extended forecasting horizons are produced recursively by sliding the input window forward and replacing the oldest historical measurements with model-generated forecasts \eqref{eq:combined_sequence}. Training is performed for 200 epochs with a batch size of 400. The Adam optimizer is used with a learning rate of \(10^{-4}\), a weight decay factor of \(10^{-5}\), and gradient clipping with a maximum norm of 1.0 to prevent divergence. Early stopping is applied with a patience threshold of 10 epochs and a minimum improvement tolerance of \(10^{-5}\). The weighting coefficients used in the training loss function \eqref{eq:total_loss}, namely \(\alpha = 0.9\) and \(\beta = 1.2\), are selected to place greater emphasis on modeling the uncertainty of residual PRBs, which is the primary objective of this work. In the results that follow, we plot the normalized residual PRB ratio. Following common practice for operational network datasets, residual PRB values are normalized to avoid disclosing absolute spectrum utilization levels.

\begin{figure*}[t] 
    \centering
    \begin{subfigure}[b]{0.45\textwidth} 
        \centering
        \includegraphics[width=\textwidth]{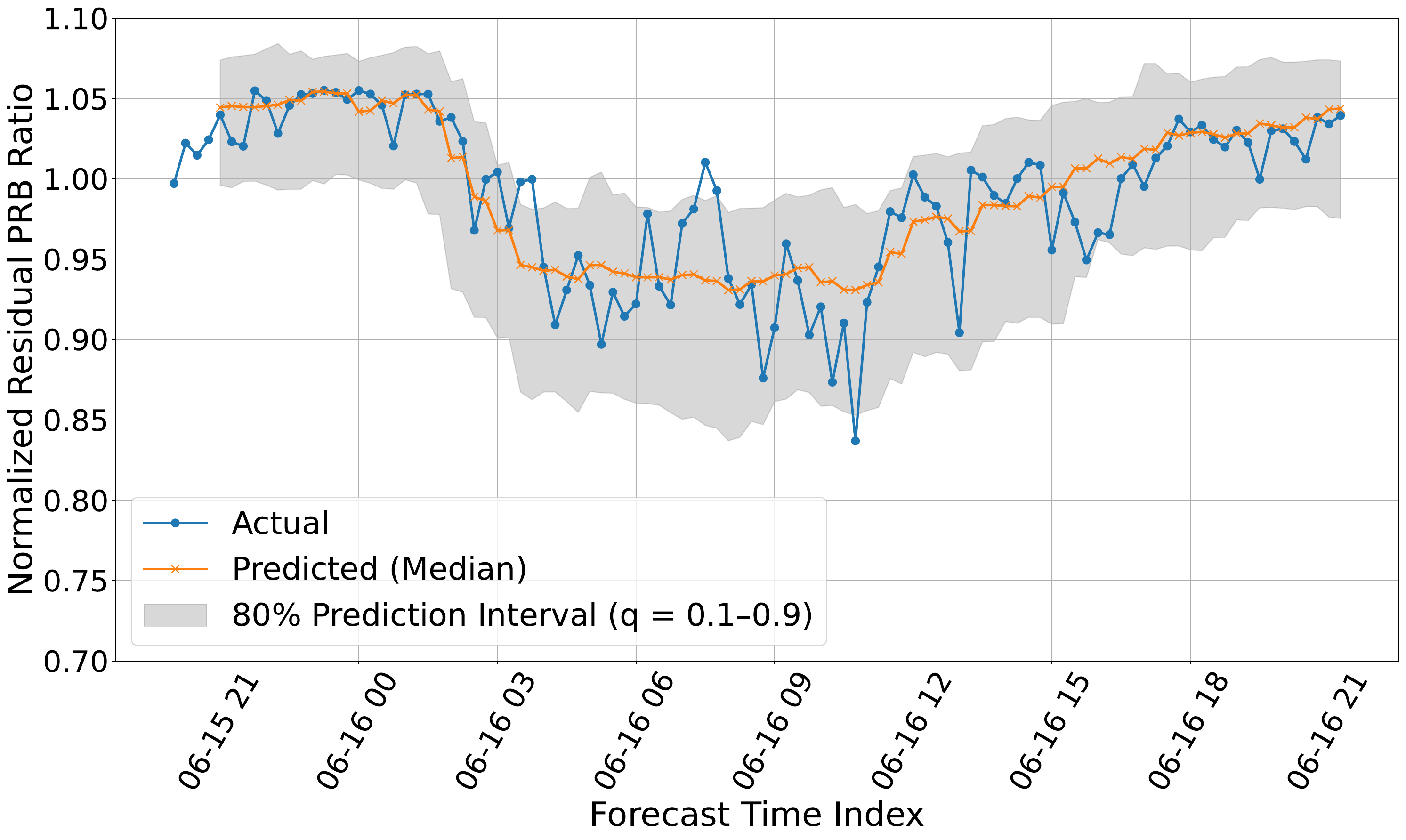}
        \caption{One day ahead residual PRB prediction at Location A.}
        \label{LocAD1}
    \end{subfigure}
    \hspace{1.5mm}
    \begin{subfigure}[b]{0.45\textwidth} 
        \centering
        \includegraphics[width=\textwidth]{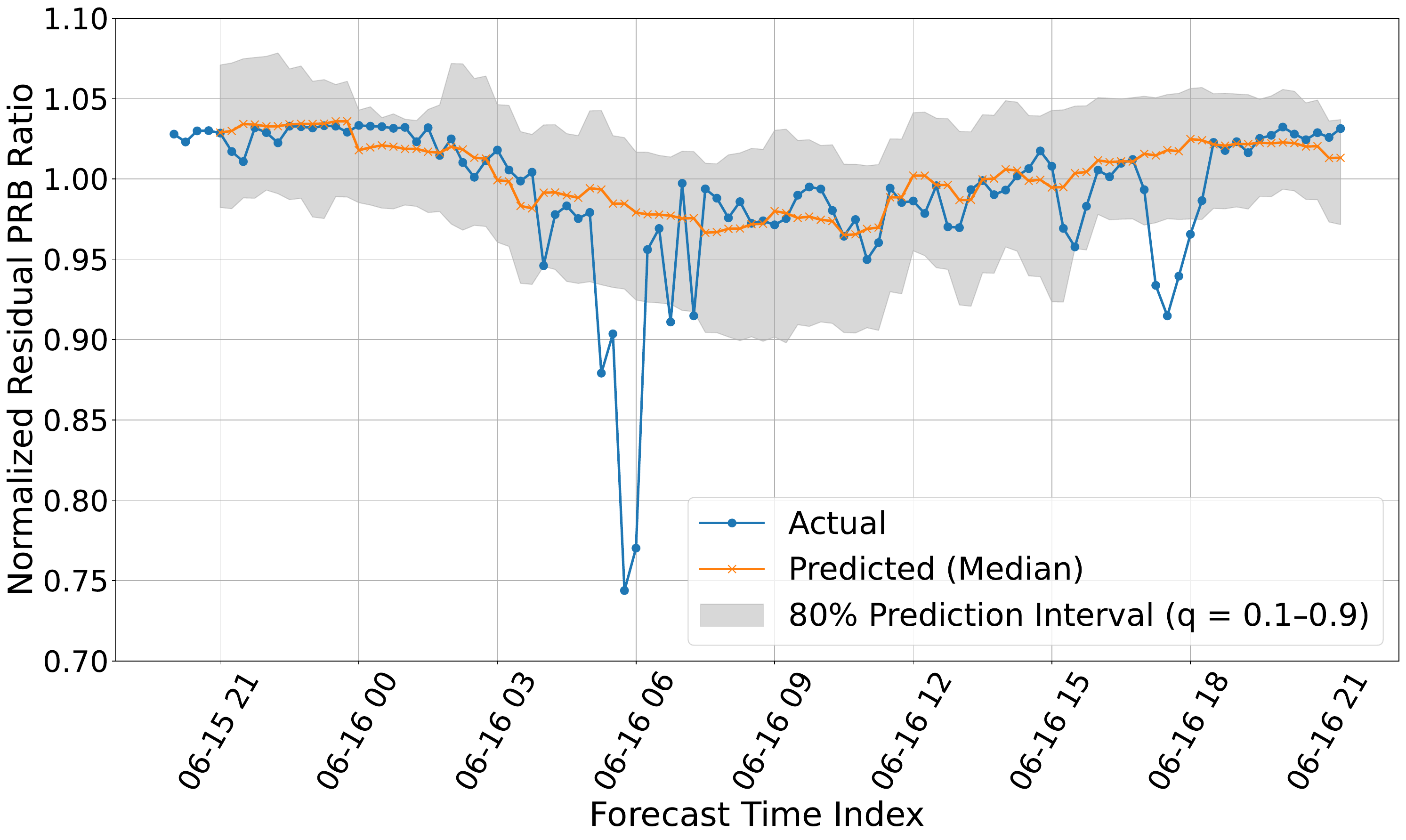}
        \caption{One day ahead residual PRB prediction at Location B.}
        \label{LocBD1}
    \end{subfigure}

    \caption{One day ahead recursive forecasting of residual PRBs for two representative LTE carriers deployed at two distinct eNB locations. }
    \label{fig:D1}
    \vspace{-5mm} 
\end{figure*}


\subsection{Evaluation Metrics}

To assess the performance of \texttt{PRB-RUPFormer} in both short- and long-horizon forecasting, two complementary classes of metrics are considered: (i) deterministic metrics that evaluate the accuracy of the median point forecast, and (ii) probabilistic calibration metrics that measure the reliability of the predicted quantile intervals. Together, these metrics quantify how well the model captures both the central trend and the uncertainty structure of residual PRBs under realistic network variability.

\subsubsection{Median Forecast Error}

The recursive forecasting pipeline relies on the median quantile (q = 0.5) for rollout, making its accuracy central to overall prediction quality. Let \(r_{t+k}\) denote the ground-truth residual PRB ratio at step \(k\), and \(\widehat{r}_{t+k,0.5}\) the corresponding predicted median. The mean absolute error (MAE) is computed as
\begin{equation}
\mathrm{MAE} = \frac{1}{K} \sum_{k=1}^{K} 
\left| r_{t+k} - \widehat{r}_{t+k,0.5} \right|~,
\label{eq:mae}
\end{equation}
where \(K\) is the forecast horizon. This metric provides a direct measure of the mismatch between the predicted and actual median trajectories.

\subsubsection{Hit Probability}

Since \texttt{PRB-RUPFormer} predicts quantiles at \(q = 0.1, 0.5,\) and \(0.9\), an \(80\) percent prediction interval is defined as
\begin{equation}
\big[ \widehat{r}_{t+k,0.1},\ \widehat{r}_{t+k,0.9} \big]~.
\label{eq:prediction_interval}
\end{equation}
The hit probability measures the proportion of time steps for which the true value lies within the predicted interval:
\begin{equation}
\mathrm{HitProb} = \frac{1}{K} \sum_{k=1}^{K} 
\mathbf{1}\!\left\{ \widehat{r}_{t+k,0.1} \le r_{t+k} \le \widehat{r}_{t+k,0.9} \right\}~.
\label{eq:hitprob}
\end{equation}
A hit probability close to the nominal confidence level (e.g., \(0.8\)) indicates well-calibrated uncertainty estimates. This metric is particularly important for network functions that depend on forecast reliability, such as proactive scheduling or slice resource guarantees.

\subsection{Experimental Results}

\subsubsection{One Day Recursive Forecasting Evaluation}


This subsection evaluates the one-day recursive forecasting performance of \texttt{PRB-RUPFormer} using an input window of \(N = 4\) historical KPI samples. The model recursively predicts the next \(M = 96\) time steps, corresponding to a full 24-hour horizon, without access to future ground-truth values. Fig. \ref{fig:D1} presents the resulting one-day ahead residual PRB forecasts for a representative LTE carrier at two geographically distinct eNB locations in the United States, referred to as Location A and Location B. For Location A, the proposed \texttt{PRB-RUPFormer} achieves a low median forecast error, with a MAE of 0.0182 on the residual PRB ratio. The predicted median trajectory closely follows the ground truth across both low-load nighttime periods and higher-variability daytime office hours. Forecast stability is reflected in a low standard deviation of the absolute median forecast error, equal to 0.0178. In addition, the hit probability reaches 0.949, significantly exceeding the nominal 0.80 confidence level, indicating that the predicted 10th–90th percentile interval reliably captures the true residual PRB values even during rapid load changes.

In contrast, Location B exhibits more pronounced short-term volatility, including sharper drops in residual PRBs during peak hours, indicative of sudden traffic bursts or scheduling irregularities. Despite these challenges, the model maintains strong predictive performance, achieving an MAE of 0.0204. The increased variability in traffic conditions is reflected in a higher standard deviation of the absolute median forecast error, equal to 0.0351, while the predicted uncertainty band appropriately expands to accommodate this variability. The corresponding hit probability of 0.898 remains above the nominal confidence level, confirming that the probabilistic forecasts remain well calibrated under more dynamic load patterns.

Overall, these results highlight two important characteristics of the proposed model. First, the median forecasts provide accurate point estimates across diverse operating regimes, as evidenced by consistently low MAE values at both locations. Second, the adaptive width of the predicted confidence intervals enables robust uncertainty quantification, maintaining high hit probabilities even in the presence of abrupt traffic fluctuations. Such behavior is particularly desirable for RAN control applications, where conservative yet informative estimates of future PRB availability are required to support proactive scheduling, congestion avoidance, and slice-level decision making.


\subsubsection{Seven Day Recursive Forecasting Evaluation}
 \begin{figure*}[t] 
    \centering
    \begin{subfigure}[b]{0.45\textwidth} 
        \centering
        \includegraphics[width=\textwidth]{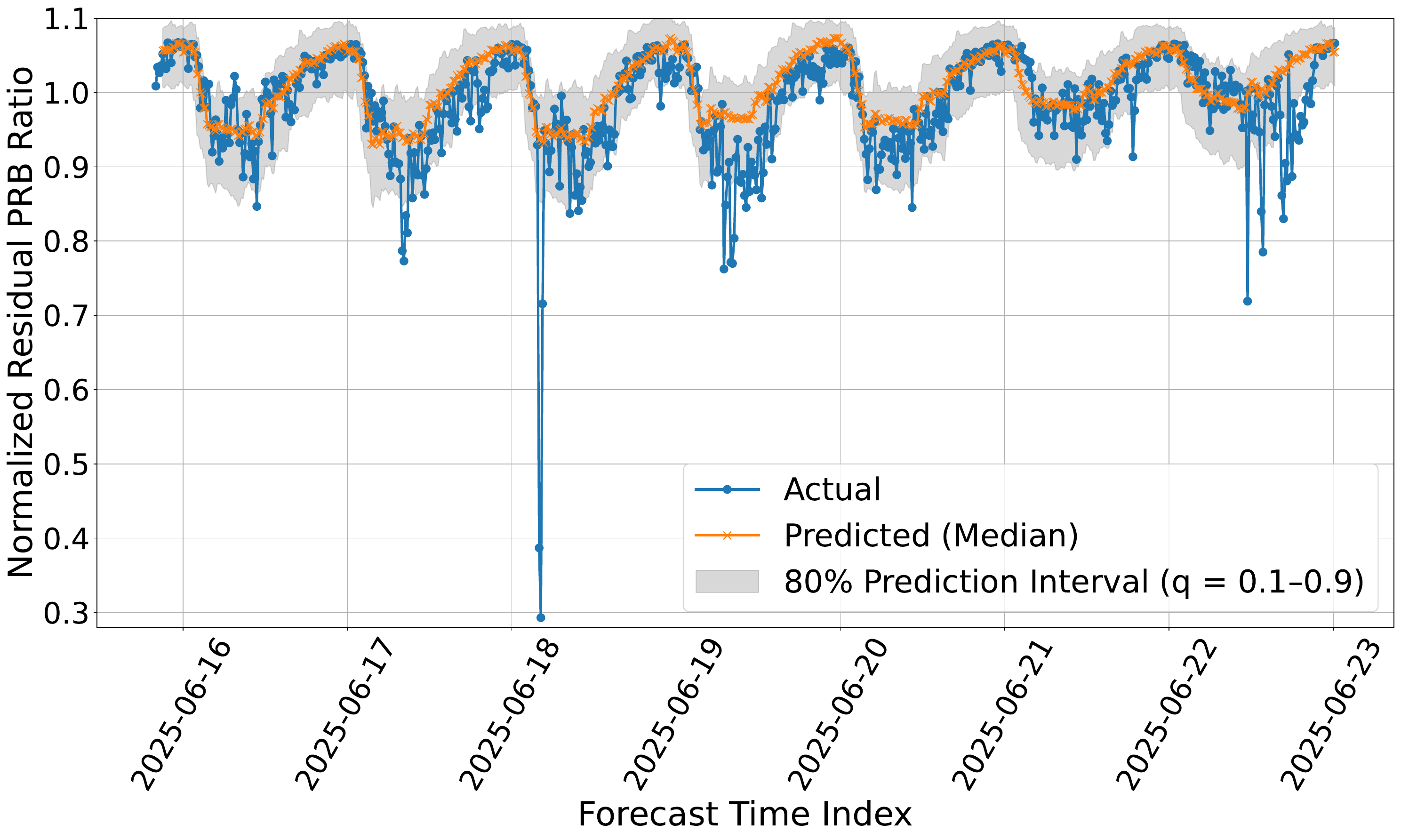}
        \caption{Seven day ahead residual PRB prediction at Location A.}
        \label{LocAD7}
    \end{subfigure}
    \hspace{1.5mm}
    \begin{subfigure}[b]{0.45\textwidth} 
        \centering
        \includegraphics[width=\textwidth]{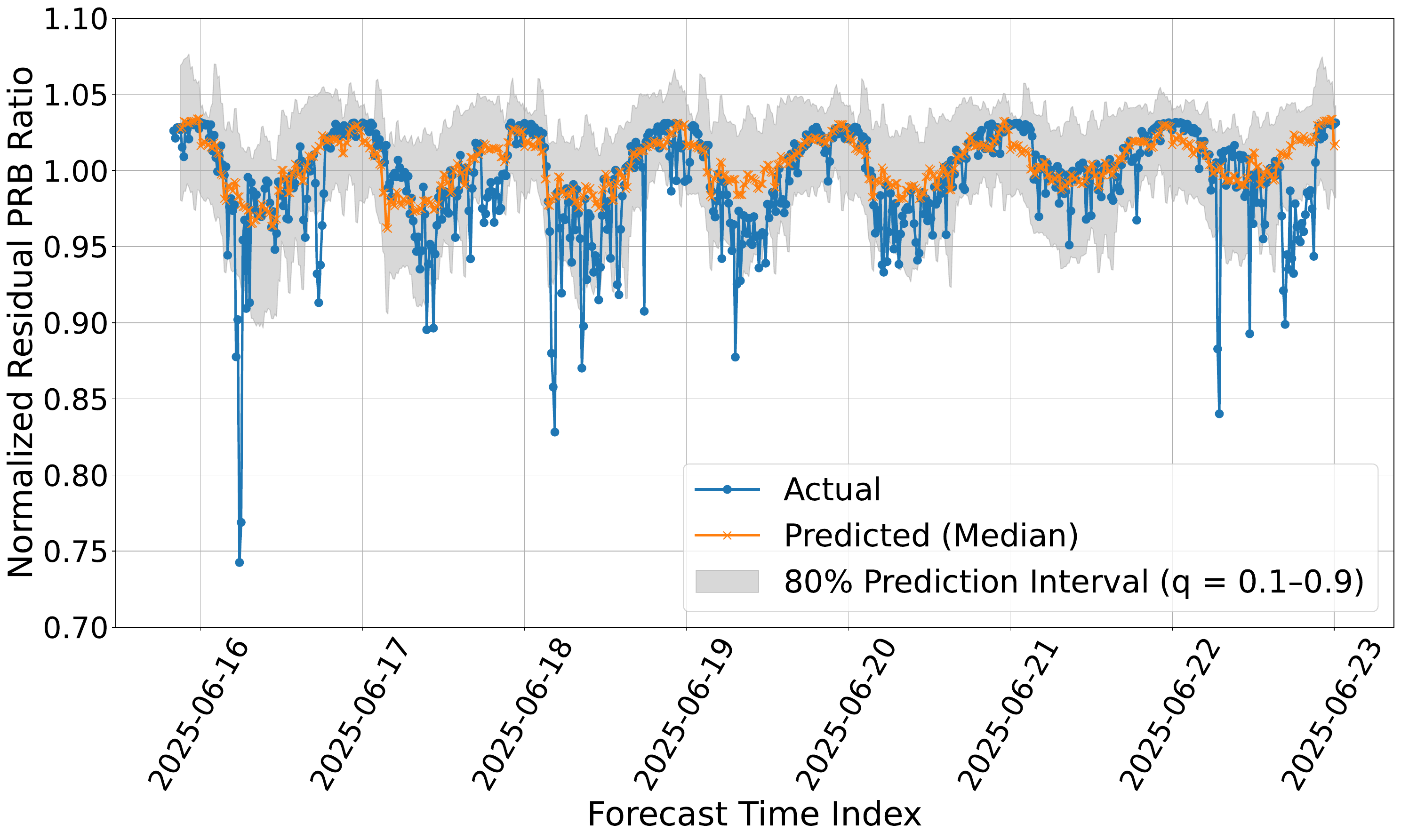}
        \caption{Seven day ahead residual PRB prediction at Location B.}
        \label{LocBD7}
    \end{subfigure}

    \caption{Seven day ahead recursive forecasting of residual PRBs for two representative LTE carriers deployed at two distinct eNB locations.}
    \label{fig:D7}
    \vspace{-5mm} 
\end{figure*}

Next, we evaluate long-horizon performance by recursively forecasting residual PRBs over a seven-day horizon. In this setting, the model repeatedly rolls forward its predictions to form subsequent inputs, thereby testing robustness to error accumulation and nonstationary traffic dynamics across multiple diurnal cycles. Fig.~\ref{fig:D7} reports results for the same two representative LTE carriers at Location A and Location B, where the median forecast (q = 0.5) is shown together with the 80\% prediction interval (q = 0.1--0.9).

At Location A, the model tracks the repeating daily structure while remaining responsive to gradual level shifts across the week. The resulting long-horizon median accuracy yields $\mathrm{MAE}=0.0298$ with $\mathrm{HitProb}=0.8863$, indicating that the predicted 10th--90th percentile band contains the ground truth substantially more often than the nominal 0.80 level. The wider uncertainty band observed during pronounced troughs reflects increased variability in residual PRB availability, which can arise from bursty traffic, scheduling changes, and short-term congestion episodes. Importantly, most sharp downward excursions remain covered by the prediction interval, suggesting that the quantile outputs capture the elevated uncertainty encountered during low-availability periods.

For Location B, the model achieves a smaller median error of $\mathrm{MAE}=0.0175$ with $\mathrm{HitProb}=0.8994$ over the full seven-day rollout. As seen in Fig.~\ref{fig:D7}(b), the predicted median closely follows the day-to-day cycles, while the prediction interval remains tight yet still encloses most abrupt deviations. Overall, both locations maintain hit probabilities above the nominal 80\% target, demonstrating that \texttt{PRB-RUPFormer} preserves probabilistic calibration under extended recursive forecasting. The observed differences between locations primarily reflect site-specific traffic volatility and the extent of abrupt residual PRB drops, which directly influence the required prediction-interval width and the difficulty of maintaining a low-error median trajectory over long horizons.

\subsubsection{Aggregate Carrier-Level Performance}
In addition to the representative carrier results shown above, we also evaluate the 7-day recursive forecasting performance across all carriers deployed at Location A. Averaging over all source cells, the proposed \texttt{PRB-RUPFormer} achieves a mean absolute error of 0.0449, with a standard deviation of 0.0207, indicating consistent prediction accuracy across heterogeneous carriers. The corresponding average hit probability is 0.8174, which remains above the nominal 0.80 confidence level. These results demonstrate that the model generalizes well beyond individual carriers and maintains well-calibrated uncertainty estimates under diverse traffic and load conditions within the same eNB.


\subsubsection{Computational Complexity and Runtime Performance}

\begin{table}[t]
    \centering
    \caption{Computational characteristics of \texttt{PRB-RUPFormer}.}
    \label{tab:complexity_short}
    \begin{tabular}{lcc}
        \toprule
        \textbf{Feature} & \textbf{GPU (V100)} & \textbf{CPU (88 threads)} \\
        \midrule
        Training Time    & 11.03 min & 12.54 min \\
        Validation Time  & 1.09 min & 1.18 min \\
        Inference Time   & \textbf{5 ms} & \textbf{5 ms} \\
        Utilization      & GPU: 8.5\% & CPU: 9.5\% \\
        \bottomrule
    \end{tabular}
\end{table}

The proposed \texttt{PRB-RUPFormer} is intentionally designed to be lightweight,
containing fewer than 0.309 million parameters with a total model size of 1.18 MB.
This compact architecture enables efficient training and real-time inference on
both GPU and CPU platforms.

Table~\ref{tab:complexity_short} summarizes the computational performance.
Training on an NVIDIA Tesla V100 completes in 11.03 minutes, while CPU training
on a 44-core/88-thread processor requires 12.54 minutes, demonstrating minimal
dependence on hardware acceleration. GPU memory usage remains low ($\approx$
80 MB), and
both platforms achieve an inference latency of \textbf{5 ms}, making the model
suitable for deployment in latency-sensitive RAN applications. These results show that the model is computationally efficient, fast to train,
and capable of real-time inference even on CPU-only systems, supporting
practical deployment in RAN controllers and edge computing nodes.

\section{Conclusion}

This paper presented \texttt{PRB-RUPFormer}, a recursive unified probabilistic Transformer for residual PRB forecasting across short- and long-term horizons in commercial LTE networks. The proposed framework jointly models multivariate KPI dynamics across all carriers and sectors of an eNB using temporal, seasonal, and carrier-aware embeddings, enabling a single lightweight model to capture cross-carrier and cross-sector coupling. By producing quantile-based prediction intervals and supporting recursive multi-step rollout, \texttt{PRB-RUPFormer} provides uncertainty-aware residual PRB forecasts over horizons ranging from minutes to multiple days. Evaluation on six months of operational LTE data from multiple U.S. locations demonstrates accurate median prediction performance and well-calibrated uncertainty estimates for both one-day and seven-day recursive forecasting, while maintaining low inference latency suitable for real-time deployment. These probabilistic residual PRB forecasts serve as a practical proxy for short- and medium-term spectrum availability and directly support spectrum-aware RAN functions such as dynamic carrier activation, congestion avoidance, and risk-aware spectrum sharing. Future work will extend the framework to multi-eNB and multi-operator scenarios, incorporate external context signals such as events and mobility indicators to improve robustness under abrupt non-stationarity, and integrate PRB forecasts into closed-loop DSA control policies for proactive spectrum allocation and sharing.


\section*{Acknowledgment}
The authors would like to thank Shankaranarayanan Puzhavakath Narayanan for insightful discussions and thoughtful comments that helped improve this work.

\bibliographystyle{IEEEtran}
\bibliography{mybib}
\end{document}